\PassOptionsToPackage{hypertexnames=false}{hyperref} 
\documentclass[11pt,a4paper]{article}

\usepackage[margin=2.5cm]{geometry}
\usepackage[authoryear,longnamesfirst]{natbib}
\usepackage{graphicx}
\usepackage[hidelinks]{hyperref}
\usepackage{fancyhdr}
\usepackage[font=small]{caption}

\usepackage{amsmath}
\usepackage{amssymb}
\usepackage{booktabs}
\usepackage{multirow}
\usepackage{array}
\usepackage{makecell}
\usepackage{float}
\usepackage{CJKutf8}

\newcommand{\Description}[1]{}

\makeatletter
\newcommand{\tblcaption}{\def\@captype{table}\caption} 
\makeatother

\fancypagestyle{plain}{\fancyhf{}\fancyfoot[C]{\thepage}}

\title{Combining Synthetic and Real Data for Low-Resource Historical OCR: A Manchu Case Study}

\author{%
  Yan Hon Michael Chung\thanks{Corresponding author. Email: hmichung@ust.hk}\\
  \small Division of Humanities, The Hong Kong University of Science and Technology,\\
  \small Clear Water Bay, Hong Kong SAR
  \and
  Hanlin Wang\thanks{Email: hl.wang@connect.ust.hk}\\
  \small The Hong Kong University of Science and Technology,\\
  \small Clear Water Bay, Hong Kong SAR
}
\date{}

\begin{document}
\renewcommand{\topfraction}{.95}
\renewcommand{\bottomfraction}{.6}
\renewcommand{\textfraction}{.03}
\renewcommand{\floatpagefraction}{.8}
\setcounter{topnumber}{3}
\setcounter{totalnumber}{5}

\maketitle

\begin{abstract}
\noindent
Manchu, now critically endangered, was one of the principal languages
of the Qing empire (1636--1912), and its extensive archival record is
increasingly digitized but remains difficult to search and analyze at
scale.
Previous work showed that vision--language models (VLMs) trained
only on synthetic Manchu word images can reach 87.4\% word accuracy
on real Qing manuscripts and prints, leaving a substantial
synthetic-to-real gap. This study examines how synthetic and real
historical training data should be combined for low-resource OCR.
Using 60{,}000 synthetic and 20{,}306 real historical word images,
we evaluate three pretrained VLMs and a compact convolutional
recurrent neural network (CRNN) under four regimes ---
synthetic-only, real-only, joint synthetic--real, and sequential
synthetic-to-real training --- under a common checkpoint-selection
and archival evaluation protocol. Introducing real training images raises the
leading configurations to 95.09--96.28\% word accuracy, while no
synthetic-only configuration exceeds 87.92\%. Synthetic
supplementation substantially improves all three VLMs, whereas its
marginal effect for the CRNN is sensitive to the training objective.
Joint and sequential training yield broadly similar archival accuracy
under the tested practical pipelines. A compact CRNN also reaches the
leading performance range once real images are available, showing
that model scale alone does not determine recognition accuracy.
Finally, complementary errors among strong recognizers allow voting
to raise accuracy to 98.27\% without additional training, while an
eighteenth-century Manchu dictionary provides a principled rule for
adjudicating disagreements.

\end{abstract}

\section{Introduction}\label{sec:introduction}

Annotated training data remain a major bottleneck for optical
character recognition (OCR), particularly for low-resource languages
and historical document collections. Synthetic data offer an
attractive response because large quantities of labelled images can
be generated without costly manual
transcription~\citep{jaderberg2014synthetic}. For historical OCR,
however, synthetic images cannot fully reproduce the typography,
handwriting variation, degradation, and capture conditions of real
documents, leaving a persistent synthetic-to-real
gap~\citep{springmann2017ocr}. Once annotated real images become
available, a practical question therefore arises. How should they be
used alongside synthetic training data?

Previous research on combining the two sources, whether through
sequential synthetic-to-real fine-tuning or joint training, has
largely concerned conventional OCR and scene-text architectures
(\citealp{martinek2019training}; \citealp{Baek_2021_CVPR};
Section~\ref{sec:lit_hybrid}). The growing use of pretrained
vision--language models (VLMs) for OCR raises the question of
whether the effects of training-data composition remain consistent
across different recognition models, especially in low-resource
settings.

Manchu provides a historically significant low-resource setting in
which to examine this question. Manchu, now critically
endangered~\citep{unesco2010atlas}, was one of the principal
languages of the Qing, the last dynasty of imperial China. The Qing
court left behind a colossal body of Manchu-language
records~\citep{crossley1993, Rawski_1996, elliott2001}, and
libraries and archives worldwide have been digitizing them. Yet much
of this material remains accessible only as scanned images rather
than machine-readable text~\citep{sun_etal_2025}.
Reliable OCR is therefore important for making these collections
searchable, indexable, and available for computational study.

Recent work demonstrates both the promise and the limitations of
synthetic training in this setting. \citet{chung2025manchu}
fine-tuned pretrained VLMs on 60{,}000 synthetic Manchu word images
without using real historical images for parameter training and
evaluated them on a 753-image benchmark drawn from Qing manuscripts
and printed sources. Their best VLM reached 87.4\% word accuracy,
compared with 56.8\% for a CRNN trained on the same synthetic corpus.
The experiment demonstrated substantial transfer from synthetic
training data to real archival images, but also left an error rate
of roughly one word in eight.

The present study examines what happens when real historical training
images are added to this setting. We combine the existing
60{,}000-image synthetic corpus with 20{,}306 annotated real Manchu
word images and evaluate three pretrained VLMs and a compact
task-specific CRNN under four training regimes. These comprise
synthetic-only, real-only, joint synthetic--real, and sequential
synthetic-to-real training, yielding 16 model $\times$ training-regime
configurations evaluated under a common checkpoint-selection and
archival evaluation protocol
(Sections~\ref{sec:dataset}--\ref{sec:methodology}). The experiment
addresses three questions. First, how much does real-image
supervision improve archival recognition relative to synthetic-only
training? Second, how do joint and sequential synthetic--real training
compare under the tested practical pipelines? Third, how does the
effect of training-data composition vary across recognition models?

The results show that real historical training data substantially
change the performance landscape. Every configuration in the leading
band uses real training images, reaching 95.09--96.28\% word
accuracy, while no synthetic-only configuration exceeds 87.92\%.
Synthetic supplementation substantially improves all three VLMs,
whereas its marginal effect for the CRNN is sensitive to the training
objective. Joint and sequential training produce broadly similar
archival accuracy under the tested pipelines, and a compact CRNN
reaches the leading performance range once real images are available
(Section~\ref{sec:results}). Finally, the strongest recognizers make
largely complementary errors. Voting over the three model-best
configurations raises archival word accuracy to 98.27\% without
additional training, with disagreements adjudicated by attestation
in an eighteenth-century Manchu
dictionary~\citep{qwj1771}
(Section~\ref{sec:results_ensemble}).

\section{Background and Related Work}\label{sec:related_work}

\subsection{Manchu Sources and Digital Access}\label{sec:lit_qing}
The Qing dynasty produced a vast body of Manchu-language records that is distinct from the parallel Chinese archive. Some Manchu documents contain information absent from Chinese-language sources, while entire classes of Qing records survive only in Manchu~\citep[pp.~74, 89]{crossley1993}. From the late twentieth century, historians increasingly emphasized the independent evidentiary value of these materials and the limitations of reconstructing Qing history from Chinese-language sources alone. Their growing use contributed to a broader reassessment of the Qing state, particularly its political, military, and frontier dimensions (\citealp{crossley1993}; \citealp[pp.~829--830]{Rawski_1996}).

Manchu is a Tungusic language written in a vertically oriented alphabetic script derived from the Mongolian script. Letters represent vowels and consonants, vary in form according to their position within a word, and are joined into continuous vertical units; dots, circles, and related graphic distinctions differentiate otherwise similar forms~\citep[pp.~13, 21--27]{li2000manchu}. These features, together with variation in handwriting, print styles, document condition, and image quality, pose distinctive challenges for OCR. Manchu is also commonly transliterated into Latin characters for scholarly use. This study follows the modified M\"{o}llendorff system presented by Roth Li~\citep[p.~16]{li2000manchu}, using romanization where necessary for lexical lookup and comparison while evaluating OCR against the original Manchu-script ground truth.

Access to Manchu sources has expanded substantially through digitization in recent decades. Major collections are available through institutions including the Biblioth\`{e}que nationale de France, the Harvard-Yenching Library, the National Palace Museum in Taipei, and the Staatsbibliothek zu Berlin~\citep{bnf_manchu, harvard_yenching_manchu, npm_manchu, sbb_manchu_collection}. Yet digitization has largely made these materials available as page images rather than machine-readable text. Much of this expanding corpus therefore remains difficult to search, index, or analyze at scale~\citep{sun_etal_2025}. Reliable OCR is a critical step toward making digitized Manchu collections searchable and computationally usable.

\subsection{Manchu OCR}\label{sec:lit_ocr}
Manchu OCR has progressed through several methodological stages. Early work beginning in the mid-2000s treated recognition primarily as a character- or sub-character-level problem, decomposing Manchu characters into strokes or intermediate Manchu Character Units (MCUs) and reconstructing higher-level forms through structural matching and classification (\citealp{zhang_2004offline}; \citealp[pp.~801--805]{zhao_2006design}; \citealp[pp.~3339--3344]{zhang_2006}). A second phase beginning around 2017 shifted toward segmentation-free, word-level recognition based on convolutional neural networks (CNNs)~\citep[pp.~46--49]{huang_2017}. Subsequent systems reported accuracies above 90\% and in some cases approaching 99\%, although these evaluations were largely restricted to relatively small closed vocabularies (\citealp{Li_2018}; \citealp{Zheng_2018}; \citealp[pp.~5--6]{Zhang_2021}). In 2022, Zhang Zhuohui released \texttt{ManchuOCR}, which combines Manchu recognition with a script-aware synthetic-image generator and supplies the synthetic corpus used in the present study (Section~\ref{sec:data_syn})~\citep{Zhang_2022}.

Recent work has expanded both recognition architectures and the scale of authentic historical training data. \citet{Wang_2024} proposed the Visual-Language framework for Manchu Word Recognition (VLMR), combining visual and semantic representations and achieving strong performance on held-out samples from its constructed datasets. More recently, \citet{bi2026scc} constructed MW14850 from 13 historical texts, comprising 124{,}448 annotated word images representing 14{,}850 lexical units, and introduced SCC$^3$, a task-specific architecture that achieved 97.16\% word accuracy on that corpus's held-out test split. These results demonstrate the strong performance attainable when specialized recognition models are trained on large corpora of authentic historical Manchu material. Assembling such corpora remains costly; verifying MW14850 involved multi-stage checks by hundreds of trained volunteers against authoritative Manchu lexicons.

A complementary line of research has examined transfer from synthetic training data. \citet{chung2025manchu} fine-tuned pretrained vision--language models (VLMs) on 60{,}000 synthetic Manchu word images and evaluated them on a separately assembled benchmark drawn from Qing manuscripts and printed sources. Their best VLM reached 87.4\% word accuracy, demonstrating substantial synthetic-to-real transfer but also a considerable gap between synthetic training and archival recognition.

Existing Manchu OCR research has therefore demonstrated both the strong performance attainable with large authentic training corpora and the potential of synthetic-to-real transfer, but the interaction between these two sources of supervision remains unclear. The present study lies between these two approaches. Rather than relying on a real-image corpus comparable in scale to MW14850, we examine whether synthetic data can be combined effectively with a smaller amount of authentic historical data, and how the resulting models transfer to a separately assembled archival benchmark spanning different Manchu historical documents.

\subsection{Synthetic--Real Training for OCR}\label{sec:lit_hybrid}
Synthetic data are widely used to address the annotation bottleneck in OCR and scene-text recognition. Large quantities of labelled text images can be generated without manual transcription, allowing recognizers to be trained when annotated real images are scarce or costly to construct~\citep{jaderberg2014synthetic}. For historical OCR, however, synthetic images cannot fully reproduce the typography, degradation, handwriting variation, and acquisition conditions of real documents, leaving a persistent synthetic-to-real gap~\citep{springmann2017ocr}. Synthetic generation therefore remains particularly useful in low-resource settings, where large annotated historical corpora may be expensive and labor-intensive to construct.

Previous studies have examined several ways of combining synthetic and real supervision. \citet{martinek2019training} compared synthetic-only, real-only, and sequential synthetic pretraining followed by real-data adaptation for historical OCR, finding synthetic-only training insufficient and sequential adaptation effective. \citet{martinek2019hybrid} improved the synthetic source itself by compositing real glyph images into generated lines before real-data fine-tuning. In scene-text recognition, \citet{Baek_2021_CVPR} directly compared joint synthetic--real training with sequential synthetic-to-real fine-tuning and found a modest advantage for sequential training in the conventional architectures they tested. These studies establish that real supervision can substantially improve synthetic-only OCR and that both joint and sequential strategies are viable.

More recently, \citet{angleraud2026structure} extended this question to VLM-based historical OCR, comparing synthetic-only, real-only, and sequential synthetic-to-real fine-tuning across several recognizers for Ancient Greek critical editions. The preferred training regime differs across models, but their experiment does not include joint synthetic--real training and does not apply the same set of regimes to task-specific OCR baselines. Previous studies have not systematically compared synthetic-only, real-only, joint synthetic--real, and sequential synthetic-to-real training across pretrained VLMs and a task-specific OCR recognizer under a common historical evaluation. The present study undertakes this comparison for Manchu, testing whether the effect of training-data composition is consistent across recognition models.

\subsection{Recognizer Voting and Lexical Post-Correction}\label{sec:lit_ensemble}
Combining predictions from multiple recognizers can improve OCR when their errors are complementary. \citet{lund2009improving} aligned the outputs of several OCR engines over degraded nineteenth-century documents and trained a selection model over the aligned hypotheses, reducing error below the best single engine. \citet{drobac2020ocr} evaluated confidence voting across neural OCR models for historical Finnish and Swedish newspapers and found that appropriate combinations further reduced recognition error. \citet{skelbye2021ocr} similarly combined five cross-fold-trained CNN--LSTM models through confidence voting for nineteenth-century Swedish newspapers, outperforming any single-model configuration. These studies show that disagreement among recognizers can provide useful evidence for selecting a more reliable transcription.

Lexical information provides another source of evidence. Dictionary-constrained recognition has a long history in handwriting recognition~\citep{koerich2003large}, while lexicon-based post-correction has been applied to low-resource languages and historical documents~\citep{kolak2005ocr, reffle2013unsupervised, nguyen2021survey}. Lexical information has also been incorporated directly into neural recognition and decoding~\citep{Nguyen_2021_CVPR}, while \citet{rijhwani2021lexically} use lexically informed correction for endangered-language OCR. Manchu is well suited to this approach because annotated historical images are scarce while extensive Qing-period dictionaries survive. The present study keeps lexical knowledge external to the recognizers. Independently trained models generate candidate transcriptions, and a period dictionary is used only to adjudicate disagreements. No additional model is trained and no new candidate transcription is generated.

\section{Dataset}\label{sec:dataset}

\subsection{Data Overview}\label{sec:data_overview}
The experiments use synthetic and real historical Manchu word images
for training and validation, together with a separate archival test
set. Table~\ref{tab:datasets} summarizes the five splits and their
roles. Training-regime labels refer to the images used for parameter
optimization; all configurations use SCI-val for checkpoint selection
(Section~\ref{sec:methodology_selection}).

\begin{table}[htbp]
\caption{The five data splits and their roles in the experimental
design.}\label{tab:datasets}
\centering\small
\begin{tabular}{llrl}
\toprule
\textbf{Split} & \textbf{Role} & \textbf{Images} & \textbf{Source} \\
\midrule
SYN-train & Synthetic training & $60{,}000$ & \texttt{ManchuOCR} pipeline~\citep{Zhang_2022} \\
SYN-val & Synthetic-domain diagnostic & $15{,}000$ & \texttt{ManchuOCR} pipeline~\citep{Zhang_2022} \\
SCI-train & Real training & $20{,}306$ & SCI-DB corpus~\citep{scidb_manchu} \\
SCI-val & Checkpoint selection & $3{,}359$ & SCI-DB corpus~\citep{scidb_manchu} \\
ARCH-test & Final archival evaluation & $753$ & \citet{chung2025manchu}; seven Qing sources \\
\bottomrule
\end{tabular}
\end{table}

\subsection{Synthetic Training Corpus}\label{sec:data_syn}
The synthetic corpus originates from \texttt{ManchuOCR}, which
generates word-level images from a lexicon of 130{,}917 Manchu entries
using multiple Manchu typefaces~\citep{Zhang_2022}. The original
release contains 750{,}000 training images and 25{,}000 evaluation
images. \citet{chung2025manchu} curated a 75{,}000-image subset
comprising 60{,}000 training images and 15{,}000 validation images,
referred to here as SYN-train and SYN-val. We use this subset
unchanged.

\begin{figure}[htbp]
\centering
\includegraphics[width=\linewidth]{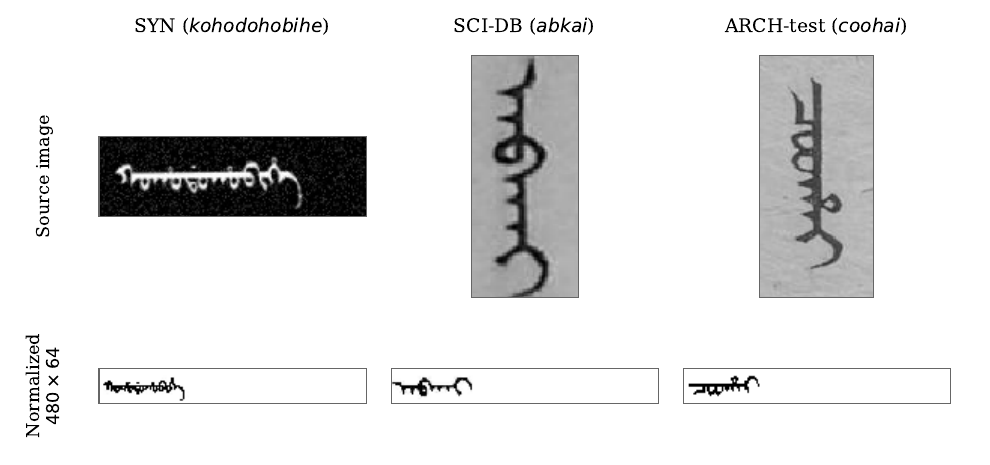}
\caption{Word images from each corpus before (top) and after (bottom)
normalization to the $480\times64$ input canvas. Synthetic words are
rendered horizontally; real word crops are vertical and are rotated
during normalization.}\label{fig:data_examples}
\end{figure}

The images are normalized to a $480\times64$ input canvas through
inversion to black-on-white, denoising, contrast enhancement, and
resizing. Figure~\ref{fig:data_examples} shows an example before and
after normalization.

\subsection{Real Historical Corpus}\label{sec:data_real}
The real-image corpus comes from SCI-DB, which contains 24{,}280 word
images extracted from Manchu books printed between 1733 and 1867,
drawn from the National Library of China's Series of Rare Ancient
Books in Manchu and Chinese~\citep{scidb_manchu}. The corpus contains 2{,}428 unique
Manchu words, each originally represented by ten image samples
scanned at 600~dpi. In the original release, word regions were
identified computationally and manually verified, and severely
degraded samples were excluded. SCI-DB was compiled by the same
research group and draws on the same National Library of China
series as the substantially larger MW14850 corpus introduced by
\citet{bi2026scc}; the present study uses only SCI-DB.

For this study, the images were normalized to a $480\times64$ input
format through grayscale conversion, denoising, contrast enhancement,
$90^\circ$ counter-clockwise rotation, and padding. A further 615
unusable images were removed after normalization, leaving 23{,}665
images. We constructed the split by holding out 1--2 images for most
word labels (2{,}306) as SCI-val and assigning the remaining usable images to
SCI-train. This yielded 20{,}306 training images and 3{,}359
validation images. Every label represented in SCI-val also occurs in
SCI-train, so SCI-val evaluates different images of vocabulary
represented during training rather than unseen-vocabulary
generalization.

\subsection{Archival Test Set and Overlap Audit}\label{sec:data_arch}
ARCH-test is the 753-image benchmark introduced by
\citet{chung2025manchu}, retained unchanged to permit direct
comparison with their synthetic-only results. It contains word-level
crops from seven Qing-period sources, including five handwritten and
two printed sources. These comprise the \textit{Old Manchu Archives
from the Grand Secretariat}~\citep{mwld_2009}, the \textit{Manchu
Veritable Records of the Taizong Emperor}~\citep{qsl_taizong}, three
eighteenth- and nineteenth-century palace
memorials~\citep{gz_zhe_qianlong,gz_zhe_xianfeng,gz_zhe_guangxu}, the
\textit{General Gazetteer of the Eight Banners}~\citep{bqtz_1739}, and
an imperial edict reproduced in Meadows's \textit{Translations from
the Manchu}~\citep{meadows_1849}. \citet{chung2025manchu} extracted
the word crops and normalized them to the same $480\times64$ input
format used here.

Exact pixel hashing found no duplicate ARCH-test images in SYN-train,
SYN-val, or SCI-train. Lexical overlap nevertheless exists because
different images can represent the same Manchu word. The SCI-train
vocabulary covers 570 of the 753 ARCH-test images, or 75.7\%, while
the combined SCI-train and SYN-train vocabulary covers 627 images, or
83.3\%. SYN-train alone covers 40.4\%. ARCH-test should therefore be
understood as an archival transfer benchmark containing both
represented and unrepresented lexical forms, rather than as a strict
open-vocabulary benchmark.

\section{Experimental Design}\label{sec:methodology}

\subsection{Recognition Models}\label{sec:methodology_grid}
We evaluate four recognition models, including three pretrained VLMs
and one task-specific CRNN. The VLMs are
LLaMA-3.2-11B-Vision-Instruct~\citep{grattafiori2024llama},
Pixtral-12B-2409~\citep{agrawal2024pixtral}, and
Qwen3-VL-8B-Instruct~\citep{qwen3vl2025}. Each is loaded through
Unsloth~\citep{unsloth2023} and fine-tuned with LoRA~\citep{hu2022lora}
on both vision and language modules using the same training recipe,
allowing the three VLMs to be compared under consistent tuning
conditions. Qwen in particular is tuned under this shared recipe
rather than a model-specific one, so its results reflect a
standardized rather than optimized configuration. Model outputs follow the Manchu-script and romanization
format used by \citet{chung2025manchu}, although all evaluation in
this study uses the Manchu-script output. Full fine-tuning and
inference settings are reported in Appendix~\ref{app:vlm_training}.

The fourth recognizer is the CRNN used by \citet{chung2025manchu},
based on the convolutional recurrent architecture of
\citet{shi2017end}. It combines a 9-layer CNN feature extractor, a
4-layer bidirectional LSTM, and a CTC output layer. The architecture
and training configuration are held fixed across the four CRNN
regimes, with full specifications reported in
Appendix~\ref{app:crnn_training}.

\subsection{Synthetic--Real Training Regimes}\label{sec:methodology_curriculum}
\begin{figure}[t]
\centering
\includegraphics[width=0.82\linewidth]{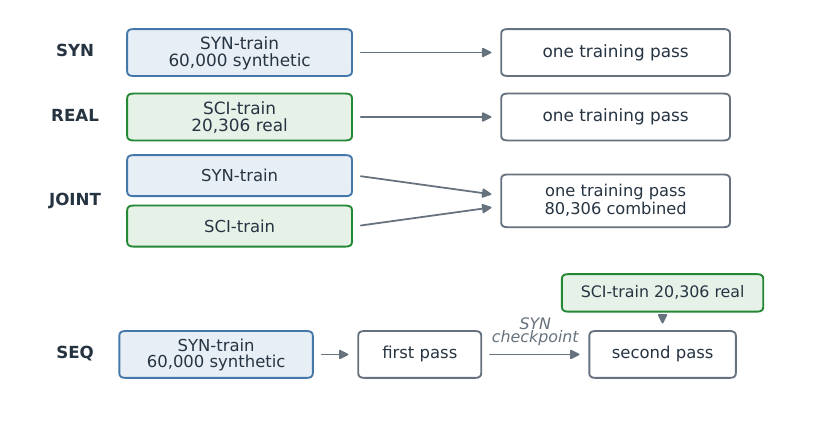}
\caption{The four training regimes. SYN and REAL use a single data
source, JOINT combines synthetic and real images in one run, and SEQ
continues real-data training from a SYN checkpoint. Blue marks synthetic
and green marks real training data. All configurations are
selected on SCI-val.}\label{fig:regime_pipeline}
\Description{Diagram of the four training regimes as data-flow arrows from the synthetic and real training corpora into one or two training passes per regime.}
\end{figure}

For each recognition model, we compare four ways of using the
synthetic and real historical training corpora shown in
Figure~\ref{fig:regime_pipeline}.

\begin{itemize}
  \item \textbf{SYN}, synthetic-only. Training uses the 60{,}000-image
        SYN-train corpus only.
  \item \textbf{REAL}, real-only. Training uses the 20{,}306-image
        SCI-train corpus only.
  \item \textbf{JOINT}, joint synthetic--real. SYN-train and SCI-train
        are combined into a single 80{,}306-image training corpus and
        shuffled together during training.
  \item \textbf{SEQ}, sequential synthetic-to-real. Training begins
        from a designated checkpoint from the corresponding SYN run
        and continues on SCI-train alone. The warm-start checkpoints
        used for each recognizer are documented in
        Table~\ref{tab:budgets}.
\end{itemize}

JOINT and SEQ therefore use the same two data sources but present
them differently. JOINT interleaves synthetic and real images within
one training run, while SEQ introduces the real images only after
synthetic training. The comparison should be interpreted as one
between the practical pipelines as realized rather than as a
controlled causal test of training order. In particular, the SEQ warm
starts differ slightly in provenance and, for LLaMA and the CRNN,
occur before the end of the corresponding SYN run. Exact checkpoint
provenance and training budgets are reported in
Table~\ref{tab:budgets}.

\subsection{Checkpoint Selection}\label{sec:methodology_selection}
For each configuration, we select the checkpoint with the highest
word accuracy on SCI-val. SYN-val is used only to measure
synthetic-domain performance, while ARCH-test is reserved for final
evaluation and plays no role in checkpoint selection.

We select the best observed checkpoint rather than the final
checkpoint because SCI-val performance is not always monotonic over
training. Ties in word accuracy are resolved first by the validation
character error rate (CER) stored in the checkpoint records (an
unstripped, macro-averaged value) and then by the earlier training
step; reported CER values are instead recomputed as the stripped,
micro-averaged measure defined in
Section~\ref{sec:methodology_metrics}. Checkpoints are saved every epoch for the CRNN and every 500
optimizer steps for the VLMs, and every saved checkpoint is
evaluated on the complete SCI-val split.

\subsection{Evaluation Metrics and Statistical Comparisons}\label{sec:methodology_metrics}
We report word accuracy (WA) and CER for each selected configuration.
All metrics are calculated from the Manchu-script output after
whitespace stripping, which removes an encoding artifact present in
the SCI-DB ground truth but not in the synthetic corpus.

Both metrics are computed over the $N$ word images of an evaluation
set, with references $g_i$ and predictions $\hat{g}_i$. WA is the
proportion of exact word matches,
\[
  \mathrm{WA} \;=\; \frac{1}{N}\sum_{i=1}^{N} \mathbf{1}\,[\hat{g}_i = g_i],
\]
and CER is the micro-averaged Levenshtein edit
distance~\citep{levenshtein1965distance} divided by total reference
length,
\[
  \mathrm{CER} \;=\; \frac{\sum_{i=1}^{N} d(\hat{g}_i, g_i)}{\sum_{i=1}^{N} |g_i|},
\]
where $d(\cdot,\cdot)$ denotes Levenshtein distance and $|g_i|$ the
character length of the reference.

For ARCH-test WA, we report 95\% non-parametric bootstrap confidence
intervals based on 1{,}000 resamples. Since all configurations are
evaluated on the same 753 images, direct comparisons use paired
statistics (Table~\ref{tab:paired_stats}), including discordant
counts and 95\% paired bootstrap intervals for the difference in WA
based on 20{,}000 resamples. These intervals resample individual word
images and should therefore be interpreted as conditional on the
fixed ARCH-test corpus rather than as estimates of generalization to
new document sources.

\section{Results}\label{sec:results}

We first examine the effect of introducing real historical training
data, then compare joint and sequential synthetic--real training, and
finally assess how the effect of training-data composition varies
across recognizers. Results are reported primarily on ARCH-test, with
complete validation and test metrics provided in
Appendix~\ref{app:full_results}.

\subsection{Effect of Real Historical Training Data}\label{sec:results_leaderboard}
\begin{figure}[t]
\centering
\includegraphics[width=0.78\linewidth]{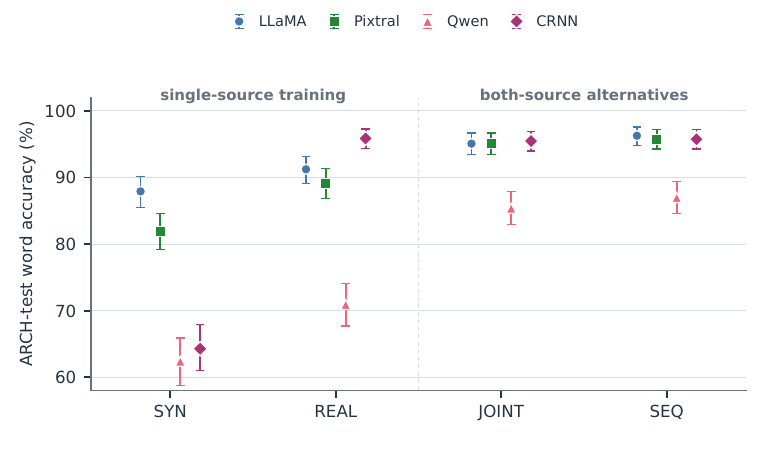}
\caption{ARCH-test word accuracy by training regime for the four
    recognition models (values of Table~\ref{tab:main_matrix}). Points show
    estimates and error bars show 95\% non-parametric bootstrap confidence
    intervals based on 1{,}000 resamples. A dashed separator distinguishes
    single-source training (SYN and REAL) from JOINT and SEQ, which are
    alternative uses of both data sources rather than successive stages.
    LLaMA and
    Pixtral lead under SYN, the CRNN leads under REAL, and three models
    converge above $95\%$ under JOINT and SEQ while Qwen remains
    lower.}\label{fig:interaction}
    \Description{Grouped point plot of ARCH-test word accuracy and bootstrap confidence intervals across four models and four categorical training regimes. A dashed separator divides single-source training (SYN and REAL) from the alternative both-source regimes (JOINT and SEQ).}
\end{figure}

Table~\ref{tab:main_matrix} reports ARCH-test WA for all 16
configurations, with Figure~\ref{fig:interaction} showing the same
results by training regime.

The strongest pattern is the effect of real historical training data.
No SYN configuration exceeds $87.92\%$ WA on ARCH-test, whereas seven
configurations using real training images fall within a leading range
of $95.09$--$96.28\%$. These are LLaMA-SEQ ($96.28\%$), CRNN-REAL
($95.88\%$), CRNN-SEQ and Pixtral-SEQ ($95.75\%$), CRNN-JOINT
($95.48\%$), and LLaMA-JOINT and Pixtral-JOINT ($95.09\%$). This
range is descriptive rather than a claim of statistical equivalence.

The synthetic-only LLaMA result also closely reproduces the earlier
benchmark. LLaMA-SYN reaches $87.92\%$, compared with $87.4\%$
reported by \citet{chung2025manchu}, despite being retrained on
different hardware and infrastructure.

\begin{table}[t]
\caption{ARCH-test word accuracy (\%) for the 16 model $\times$
training-regime configurations. Each cell is the
SCI-val-peak checkpoint selected by the rule of
Section~\ref{sec:methodology_selection}, scored on the complete
753-image ARCH-test split. Bold marks each recognition model's best
regime. Complete word-accuracy and character-error-rate results on
all three splits, with bootstrap intervals, are in
Table~\ref{tab:perf_manchu}
(Appendix~\ref{app:full_results}).}\label{tab:main_matrix}
\centering
\begin{tabular}{lrrrr}
\toprule
\textbf{Model} & \textbf{SYN} & \textbf{REAL} & \textbf{JOINT} & \textbf{SEQ} \\
\midrule
LLaMA   & 87.92 & 91.24 & 95.09 & \textbf{96.28} \\
Pixtral & 81.94 & 89.11 & 95.09 & \textbf{95.75} \\
Qwen    & 62.42 & 70.92 & 85.39 & \textbf{86.99} \\
CRNN    & 64.28 & \textbf{95.88} & 95.48 & 95.75 \\
\bottomrule
\end{tabular}
\end{table}

\subsection{Joint versus Sequential Training}\label{sec:results_curriculum}
Across all four recognition models, SEQ has a slightly higher
ARCH-test point estimate than JOINT in the main 16-configuration
grid. The differences are $0.27$~percentage points for the CRNN,
$0.66$ for Pixtral, $1.20$ for LLaMA, and $1.59$ for Qwen (computed
from the per-item counts, so the last digit can differ from
subtraction of the rounded table values). However, every paired 95\%
CI includes zero (Table~\ref{tab:paired_stats}), so these results do
not establish a consistent advantage for SEQ over JOINT.

As described in Section~\ref{sec:methodology_curriculum}, the
comparison is between the realized training pipelines, which differ
slightly in warm-start provenance and total updates, rather than a
controlled test of training order.

A clearer difference between the two regimes appears on SYN-val.
CRNN-SEQ retains $76.91\%$ WA on SYN-val, compared with $99.39\%$ for
CRNN-JOINT, while their ARCH-test performance remains similar. In
this case, sequential real-data fine-tuning substantially reduces
synthetic-domain performance without producing a corresponding
archival gain.

\subsection{Recognizer-Dependent Effects of Training Data}\label{sec:results_xfamily}
The effect of training-data composition differs substantially across
recognition models (Figure~\ref{fig:interaction}). Under SYN, LLaMA
and Pixtral lead at $87.92\%$ and $81.94\%$ WA, while the CRNN
reaches $64.28\%$\footnote{This figure is higher than the $56.8\%$
reported for the same CRNN architecture on the same 753-image test
split by \citet{chung2025manchu}. The two figures come from different
training runs on different hardware, so the gap conflates run-to-run
variation with checkpoint selection. Within our run the selection
effect is large: read at a fixed mid-training epoch the same model
scores $52.99\%$, whereas the SCI-val-peak rule of
Section~\ref{sec:methodology_selection} selects a later checkpoint
reaching $64.28\%$. The relative ordering between the synthetic-only
CRNN and the synthetic-only LLaMA and Pixtral models is unchanged.}
and Qwen $62.42\%$. Under REAL, the ordering changes markedly. The
CRNN reaches $95.88\%$, compared with $91.24\%$ for LLaMA, $89.11\%$
for Pixtral, and $70.92\%$ for Qwen. Under JOINT, LLaMA, Pixtral, and
the CRNN converge within a narrow range of $95.09$--$95.48\%$, while
Qwen remains lower at $85.39\%$.

The marginal value of synthetic supplementation once real data are
available is likewise recognizer-dependent. Moving from REAL to JOINT
raises WA by $3.85$~percentage points for LLaMA, $5.98$ for Pixtral,
and $14.48$ for Qwen. For the CRNN, JOINT is $0.40$~points below REAL
under the released training recipe, although a corrected-objective
robustness check in Appendix~\ref{app:crnn_training} shows this
fine-grained ordering is implementation-sensitive. Overall, the additional value of
synthetic data varies substantially across recognizers.

\section{Residual Error Analysis and Dictionary-Guided Voting}\label{sec:voting}

The strongest individual configuration still misrecognizes 28 of the
753 ARCH-test words. We therefore examine whether the remaining
errors are shared across recognizers and whether disagreement among
strong recognizers can be used to improve final transcription
accuracy.

\subsection{Residual Errors and Recognizer Complementarity}\label{sec:results_errors}
\begin{figure}[t]
\centering
\includegraphics[width=0.55\linewidth]{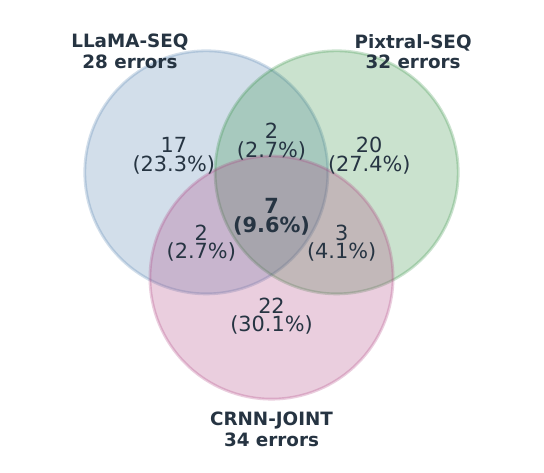}
\caption{Overlap of ARCH-test error sets for LLaMA-SEQ, Pixtral-SEQ,
and CRNN-JOINT. Counts and percentages refer to the 73-image union of
the three error sets. The Venn diagram records whether each
recognizer is incorrect, not whether their incorrect transcriptions
are identical. Of these 73 images, six receive the same incorrect
transcription from all three recognizers; the remaining 67 are
disagreement cases. Circle areas are schematic rather than
proportional to set size.}\label{fig:error_overlap}
\Description{Three-circle overlap diagram of the error sets of LLaMA-SEQ, Pixtral-SEQ, and CRNN-JOINT on the 753-image ARCH-test split.}
\end{figure}

We compare LLaMA-SEQ, Pixtral-SEQ, and CRNN-JOINT, which make 28, 32,
and 34 errors on ARCH-test, respectively. Their error sets contain 73
unique images in total (Figure~\ref{fig:error_overlap}). The
three-way intersection contains seven images on which all three
recognizers are incorrect; this intersection is defined by
correctness and does not imply identical predictions. Among the 73
images, six receive the same incorrect transcription from all three
recognizers. The remaining 67 images contain non-identical recognizer
outputs and are therefore disagreement cases.

Most residual errors are also small at the character level. Of the
incorrect transcriptions, 26 of 28 for LLaMA-SEQ, 27 of 32 for
Pixtral-SEQ, and 32 of 34 for CRNN-JOINT are at Levenshtein
distance~1 from the reference. A one-letter perturbation of
an attested Manchu word rarely yields another attested word, so most
of these near-miss transcriptions are detectable by dictionary
lookup. This combination of complementary predictions and
predominantly single-character errors motivates the lexical
adjudication procedure described in
Section~\ref{sec:methodology_ensemble}.

\subsection{Dictionary-Guided Adjudication}\label{sec:methodology_ensemble}
The ensemble combines LLaMA-SEQ, Pixtral-SEQ, and CRNN-JOINT,
selected solely on SCI-val. Qwen is excluded because its best
configuration performs more than three percentage points below the
other models on SCI-val. ARCH-test plays no role in selecting the
ensemble members.

Disagreements among their predictions are adjudicated using the
\emph{Imperially Commissioned Enlarged and Revised Mirror of the Qing
Language} (\begin{CJK}{UTF8}{bsmi}御製增訂清文鑑\end{CJK}, QWJ), an
eighteenth-century Manchu dictionary~\citep{qwj1771}. After
normalization, its digitized headwords provide an attestation lexicon
of 12{,}891 distinct word tokens. Each candidate transcription is
deterministically transliterated from Manchu script into
M\"ollendorff romanization and normalized to the same form used for
dictionary lookup. The VLM-generated romanization is not used.

\begin{figure}[t]
\centering
\includegraphics[width=0.70\linewidth]{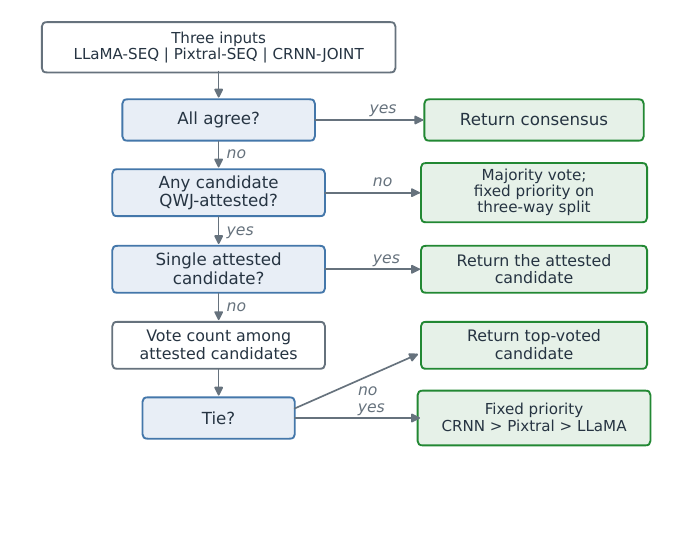}
\caption{Dictionary-guided adjudication of the three recognizer
outputs. QWJ attestation is used to resolve disagreements, with vote
count and SCI-val-based priority used for remaining
ties.}\label{fig:voting_flowchart}
\Description{Flowchart of the five-step voting rule, from consensus check through dictionary attestation, vote count, and fixed-priority tie-breaking.}
\end{figure}

The adjudication rule is shown in
Figure~\ref{fig:voting_flowchart}. If all three recognizers agree,
their consensus is returned. On disagreement, QWJ-attested candidates
are preferred. If several candidates are attested, vote count
determines the output, with a fixed priority based on SCI-val
accuracy used to break residual ties. If no candidate is attested,
the system falls back to majority voting under the same priority. The
rule has no trainable parameters.

\subsection{Voting Performance and Residual Errors}\label{sec:results_ensemble}
\begin{figure}[t]
\centering
\includegraphics[width=0.88\linewidth]{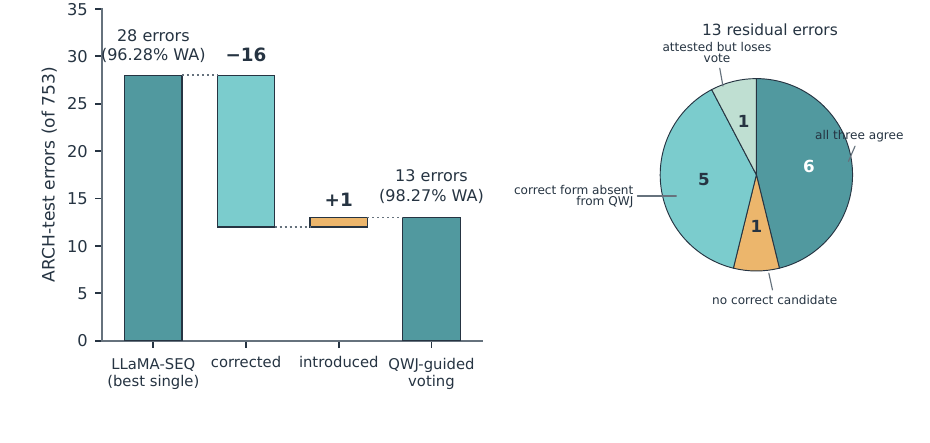}
\caption{Residual ARCH-test errors under voting adjudication.
The waterfall (left) shows 28 errors for LLaMA-SEQ, 16 corrected and one
introduced, leaving 13 under the final voting rule; the pie chart (right)
decomposes the four categories of the 13 residual errors.}
\label{fig:voting_errors}
\Description{Left, a waterfall chart shows 28 ARCH-test errors for LLaMA-SEQ, a reduction of 16 errors through voting, one introduced error, and 13 errors after QWJ-guided voting. Right, a pie chart decomposes those 13 residual errors into four categories.}
\end{figure}

Combining the three recognizers substantially reduces residual error.
Dictionary-guided voting raises ARCH-test WA from $96.28\%$ for the
best single configuration to $98.27\%$, reducing the number of errors
from 28 to 13 (Figure~\ref{fig:voting_errors}). Of the 67
disagreement cases, the adjudication rule returns the correct
transcription in 60, leaving seven unresolved. In addition, the six
cases in which all three recognizers produce the same incorrect
transcription remain incorrect. These two groups together account for
the 13 residual errors.
Relative to LLaMA-SEQ, the rule corrects 16 errors while introducing one,
with a paired 95\% CI of $[+0.93, +3.05]$~percentage points
(Table~\ref{tab:paired_stats}).

The improvement primarily reflects the complementary errors
documented in Section~\ref{sec:results_errors}. Plain majority voting
also reaches $98.27\%$ WA on ARCH-test, so QWJ guidance does not
further increase the final score for this test set. Its contribution
is instead to provide a principled adjudication rule when recognizers
disagree, particularly when all three produce different
transcriptions and no majority exists. On SCI-val, QWJ guidance
produces a small additional gain, raising WA from $99.64\%$ under
majority voting to $99.67\%$. The result is also robust to
composition choices. Substituting either of the CRNN's other leading
configurations shifts ARCH-test accuracy by at most $0.40$~points
($97.88$--$98.27\%$), and reversing the fallback priority leaves
SCI-val unchanged and moves a single ARCH-test item.

The 13 residual ARCH-test errors indicate the limits of
candidate-based lexical adjudication. In the six identical-error
cases, all three recognizers produce the same incorrect
transcription, leaving no alternative candidate to select. Among the
seven unresolved disagreement cases, one contains no correct
candidate, five contain a correct transcription absent from the QWJ
token lexicon because it is an inflected or derived form, and one
contains an attested correct transcription that loses to a competing
attested candidate under the voting and priority rule.

\section{Discussion}\label{sec:discussion}

Three findings emerge from the experiment. First, introducing real
historical training images substantially reshapes the performance
landscape, bringing several recognizers into the $95$--$96\%$ WA
range and allowing a compact CRNN to compete with much larger
pretrained VLMs. Second, the value of synthetic supplementation
differs across recognizers, while the experiments do not identify a
universal advantage for either joint or sequential training. Third,
strong recognizers retain sufficiently complementary errors that
their combination raises ARCH-test WA to $98.27\%$, with period
lexicography providing a principled mechanism for adjudicating
disagreements.

\subsection{Real Data and Model Scale}\label{sec:disc_reversal}
The clearest result is that access to real historical training
images changes both absolute performance and the relative ordering
of recognizers (Table~\ref{tab:main_matrix}). Under synthetic-only
training, LLaMA and Pixtral substantially outperform the CRNN. Once
real images are introduced, however, the CRNN reaches the same
leading performance range as the strongest VLM configurations. The
best CRNN and VLM results differ by less than half a percentage
point on ARCH-test, with the difference falling within paired
uncertainty (Table~\ref{tab:paired_stats}). Model scale alone is
therefore not a reliable predictor of recognition accuracy in this
setting.

This does not imply that CRNNs are generally preferable to VLMs.
The VLMs remain competitive recognizers, and their prompt-based
interface is directly relevant to downstream tasks such as
translation, normalization, and structured extraction. What the
result does show is that a compact task-specific recognizer can
remain highly competitive when appropriate real historical
supervision is available. The robust finding is therefore convergence
in performance, not superiority of one architecture or training
regime. For low-resource historical
OCR, improvements in training data can consequently matter at least
as much as increases in model scale.

Model compactness also has practical consequences beyond accuracy.
As logged by the evaluation harness, the $\sim$12M-parameter CRNN
transcribes a word in about $8$~ms on a single H200 GPU, while the VLMs take
$1.5$--$2.3$~s per word under 4-bit greedy decoding --- observed
end-to-end pipeline timing rather than a device-level benchmark, but
a gap of more than two orders of magnitude that lowers the
infrastructure barrier for serving and repeated batch inference
over large cultural-heritage collections.

\subsection{Recognizer-Dependent Training Effects}\label{sec:disc_interaction}
The effect of synthetic supplementation is not consistent across
recognizers. Moving from REAL to JOINT improves all three VLMs, by
$3.85$~percentage points for LLaMA, $5.98$ for Pixtral, and $14.48$
for Qwen, whereas the CRNN shows a small decline under the released
training recipe (Section~\ref{sec:results_xfamily}). The broader
conclusion is therefore not that synthetic supplementation is universally
beneficial or unnecessary, but that its value is conditional on the
recognizer and training implementation.

This qualifies conclusions drawn from earlier synthetic--real OCR
studies. Sequential synthetic pretraining followed by real-data
adaptation has proven effective for historical
OCR~\citep{martinek2019training,martinek2019hybrid}, while joint and
sequential strategies have both been successful in scene-text
recognition~\citep{Baek_2021_CVPR}. More recent work on Ancient
Greek likewise finds that the preferred training regime differs
across recognizers~\citep{angleraud2026structure}. Our results
extend this point by applying the same four training regimes across
pretrained VLMs and a task-specific CRNN. Training-data strategies
established for one recognizer class should therefore not be assumed
to transfer unchanged to another.

The comparison between JOINT and SEQ leads to a related conclusion.
Although SEQ has slightly higher point estimates for all four
recognizers in the main grid, every paired confidence interval
includes zero (Table~\ref{tab:paired_stats}). Neither schedule
therefore emerges as generally superior. In practice, the choice
between joint and sequential training may depend as much on
operational considerations --- such as the second training stage and
checkpoint hand-off that sequential training requires --- and on
model-specific behavior as on archival accuracy alone.

\subsection{Complementarity and Lexical Adjudication}\label{sec:disc_practical}
High individual accuracy does not imply that strong recognizers fail
on the same images. LLaMA-SEQ, Pixtral-SEQ, and CRNN-JOINT each
exceed $95\%$ WA, yet only seven ARCH-test images are misrecognized
by all three (Section~\ref{sec:results_errors}). Their residual
errors are therefore sufficiently complementary that combining their
outputs reduces the number of errors from 28 for the best single
recognizer to 13, raising WA from $96.28\%$ to $98.27\%$
(Section~\ref{sec:results_ensemble}). This gain requires no
additional model training and arises primarily from diversity in
recognizer errors rather than from further improvement to any
individual model.

The role of the historical dictionary is more specific. Plain
majority voting reaches the same $98.27\%$ WA on ARCH-test, so QWJ
attestation is not the source of the ensemble's overall accuracy
gain on this benchmark. Its contribution is to provide a
reproducible and historically grounded rule for disagreements in
which no majority exists or several candidates remain plausible.

This suggests a broader use for historical lexical resources in
low-resource OCR. Languages with limited annotated image data may
nevertheless possess dictionaries, glossaries, or other structured
lexical resources created within the same documentary tradition.
Such resources can be incorporated after recognition as an
adjudication layer without retraining the underlying models. Their
usefulness remains bounded by lexical coverage, as the remaining
ARCH-test errors show, but they provide a practical way to connect
historical linguistic resources with modern recognition pipelines.

Taken together, the findings suggest a practical sequence for
low-resource historical OCR. Synthetic data provide a useful
starting point when annotated historical images are unavailable, but
real historical supervision should be incorporated once it can be
obtained. The appropriate synthetic--real composition should be
evaluated for the recognizer at hand rather than inherited from
another architecture. Where residual accuracy is important,
heterogeneous recognizers and existing historical lexical resources
can provide an additional layer of error reduction without further
model training.

\section{Limitations and Future Work}\label{sec:disc_limits}
The conclusions rest on a fixed archival benchmark of 753 word
images drawn from seven Qing-period sources
(Section~\ref{sec:data_arch}). Retaining ARCH-test unchanged permits
direct comparison with \citet{chung2025manchu}, but its size limits
discrimination among the strongest configurations, which is why the
cross-model claims of Section~\ref{sec:disc_reversal} are stated as
convergence within a leading band rather than a ranking. The
benchmark also contains substantial lexical overlap with the
training data and should therefore be interpreted as an archival
transfer benchmark rather than a strict open-vocabulary test. The
reported confidence intervals are likewise conditional on this
corpus and do not quantify uncertainty in generalization to new
sources or vocabulary, a limitation that a larger and more diverse
archival test corpus would most directly address.

All real training images come from SCI-DB~\citep{scidb_manchu}.
Although ARCH-test provides evaluation on a separate set of
manuscripts and printed sources, the training side of the experiment
does not test whether the same synthetic--real patterns hold when
real supervision is drawn from multiple historical corpora with
greater variation in script style, document type, and image quality.
Extending the experiment to larger and more heterogeneous real
training corpora is therefore an important next step.
MW14850~\citep{bi2026scc} provides one natural resource for
increasing the scale of authentic supervision, while additional
collections would be needed to broaden variation in document type,
script style, and image quality.

Each configuration in the main grid is represented by a single
training run. The reported bootstrap intervals quantify variation
over evaluation items rather than stochastic variation in
initialization, data order, LoRA training, or checkpoint selection.
The corrected-objective CRNN controls further change both the
optimization objective and the realized training trajectory, so
objective sensitivity cannot be cleanly separated from run-to-run
variation. The CRNN composition and schedule conclusions are accordingly stated
for the released recipe's runs only. In addition,
JOINT and SEQ should be understood as the realized pipelines
described in Section~\ref{sec:methodology_curriculum} rather than as
a controlled isolation of training order. Replicated runs of the
leading configurations would provide a stronger basis for estimating
these effects.

The evaluation is word-level and assumes that word regions have
already been identified. The reported results therefore do not
measure page-layout analysis, segmentation, or end-to-end
transcription of complete historical documents. Extending the
workflow from word recognition to page-level OCR remains necessary
for deployment across large archival collections.

Finally, dictionary-guided adjudication depends on the coverage of
the QWJ lexicon. Since QWJ comes from the same Qing textual
tradition as the evaluation material, it provides a favorable
historical lexical resource rather than a domain-independent
linguistic prior. It cannot reliably resolve forms absent from its
token inventory, including some inflected and derived forms, and can
occasionally prefer an attested but incorrect candidate. Transfer to
genres, periods, and lexical domains not represented in the present
benchmark remains to be established. Broader morphological resources
and contextual information beyond the isolated word may further
reduce these residual errors.

\section{Conclusion}\label{sec:conclusion}
This study examined how synthetic and real historical training data
can be combined for low-resource OCR using Manchu as an empirical
case. Across four recognizers and four training regimes, the
clearest result is the importance of real historical supervision. No
synthetic-only configuration exceeds $87.92\%$ WA on ARCH-test,
while several configurations using real training images reach
$95.09$--$96.28\%$. A compact CRNN can also reach this leading
performance range, showing that model scale alone does not determine
recognition accuracy.

The effect of synthetic supplementation, however, varies across
recognizers. Adding synthetic data to real training substantially
improves all three VLMs, while the direction and magnitude of the
effect for the CRNN are sensitive to the training objective. The
experiments likewise identify no universal advantage for joint over
sequential training, or vice versa. Training-data strategies should
therefore be evaluated for the recognizer at hand rather than
assumed to transfer across model classes.

Finally, residual errors among strong recognizers are highly
complementary. Combining LLaMA-SEQ, Pixtral-SEQ, and CRNN-JOINT
raises ARCH-test WA from $96.28\%$ to $98.27\%$ without additional
model training. The gain arises primarily from recognizer
complementarity, while an eighteenth-century Manchu dictionary
provides a principled external rule for adjudicating disagreements.
Based on the results, we propose that synthetic data can bootstrap
recognition when annotation is unavailable, real historical images
should be incorporated once they can be obtained, compact
task-specific recognizers should remain in the candidate pool
alongside VLMs, and heterogeneous recognizers and existing lexical
resources can further reduce residual error.

\bibliographystyle{cas-model2-names}
\bibliography{reference}

\appendix
\setcounter{table}{0}
\setcounter{figure}{0}
\counterwithin{table}{section}
\counterwithin{figure}{section}
\section{Training Details}\label{app:training_procedure}

\subsection{VLM Training}\label{app:vlm_training}
All three VLMs share one fine-tuning and inference recipe
(Table~\ref{tab:vlm_recipe}); training budgets and the SEQ
warm-start checkpoints are given in Table~\ref{tab:budgets}.

\begin{table}[htbp]
\caption{Shared VLM fine-tuning and inference recipe (all 12 VLM cells; inference settings identical for sweep-time SCI-val scoring and final ARCH-test scoring).}\label{tab:vlm_recipe}
\centering\small
\begin{tabular}{lp{0.66\linewidth}}
\toprule
\textbf{Item} & \textbf{Setting} \\
\midrule
LoRA & rank $32$, $\alpha{=}64$, dropout $0.05$; vision + language modules \\
Optimizer & 8-bit AdamW (\texttt{paged\_adamw\_8bit}), lr $1{\times}10^{-4}$, weight decay $0.01$ \\
Schedule & cosine with warm restarts (single cycle; no restart fires), $1{,}000$ warmup steps \\
Duration / precision & 5 epochs, bfloat16 mixed precision \\
Parallelism / batch & 4 data-parallel GPUs, per-device batch 4, grad.\ accum.\ 1 (global 16) \\
Quantization & 4-bit base weights during fine-tuning (QLoRA), matching 4-bit inference \\
Random seed & $3407$ (all VLM runs) \\
Initialization & \makecell[l]{SYN/REAL/JOINT: public instruction-tuned backbones;\\ SEQ: designated checkpoint from the corresponding SYN run (Table~\ref{tab:budgets})} \\
\midrule
Inference loading & Unsloth \texttt{FastVisionModel.from\_pretrained} $\rightarrow$ \texttt{for\_inference}; 4-bit weights (\texttt{load\_in\_4bit}); \texttt{sdpa} attention \\
Decoding & greedy via the \texttt{transformers} \texttt{generate} interface; \texttt{max\_new\_tokens} $1536$; \texttt{use\_cache}; pad token = EOS; no sampling, temperature, top-$k$/top-$p$, single beam \\
Prompt & fixed instruction, identical across all 12 cells and identical to the fine-tuning prompt; single-turn chat template with generation prompting \\
Prompt text & {\itshape You are an expert OCR system for Manchu script. Extract the text from the provided image with perfect accuracy. Format your answer exactly as follows: first line with `Manchu:' followed by the Manchu script, then a new line with `Roman:' followed by the romanized transliteration.} \\
Output format & \texttt{Manchu:} line (U+1800--U+18AF) + \texttt{Roman:} line (M\"ollendorff); the CRNN emits Manchu only, so all cross-model metrics use the Manchu field \\
Parsing & newline split; case-insensitive \texttt{Manchu:}/\texttt{Roman:} prefix match; split on first colon; strip whitespace; per-sample record stores predictions, ground truth, and wall-clock inference time \\
\bottomrule
\end{tabular}
\end{table}

\begin{table}[htbp]
\caption{Training budgets (optimizer steps at fixed batch size) and
SEQ-stage warm starts. The JOINT budget equals the SYN and REAL
budgets combined for both the VLMs and the CRNN
($25{,}100 = 18{,}750 + 6{,}350$; $502{,}000 = 375{,}000 +
127{,}000$), and every training image is scheduled for the same
number of passes under either regime.}\label{tab:budgets}
\centering\small
\begin{tabular}{lrrl}
\toprule
\textbf{Regime} & \textbf{VLM steps} & \textbf{CRNN steps} & \textbf{Training data} \\
\midrule
SYN & $18{,}750$ & $375{,}000$ & $60{,}000$ synthetic \\
REAL & $6{,}350$ & $127{,}000$ & $20{,}306$ real \\
JOINT & $25{,}100$ & $502{,}000$ & $80{,}306$ combined \\
SEQ & $6{,}350$ & $127{,}000$ & $20{,}306$ real, warm-started \\
\midrule
\multicolumn{4}{l}{\textit{Warm starts}: LLaMA-SEQ $\leftarrow$ synthetic step $17{,}000/18{,}750$ (SCI-val-selected);} \\
\multicolumn{4}{l}{Pixtral-SEQ and Qwen-SEQ $\leftarrow$ step $18{,}750$ (end of training); CRNN-SEQ $\leftarrow$ step $251{,}250/375{,}000$.} \\
\multicolumn{4}{l}{All four starting points lie within $1.2$~points of their runs' full-split SCI-val peaks.} \\
\bottomrule
\end{tabular}
\end{table}

\subsection{CRNN Training}\label{app:crnn_training}
The CRNN uses the architecture and training hyperparameters of
\citet{chung2025manchu} (Tables~\ref{tab:crnn_arch}
and~\ref{tab:crnn_training}); the four cells differ only in
training-data composition.

\begin{table}[htbp]
\caption{CRNN architecture (identical across the four \texttt{crnn-*}
cells; inherited unchanged from \citealp{chung2025manchu}).}\label{tab:crnn_arch}
\centering\small
\begin{tabular}{lp{0.68\linewidth}}
\toprule
\textbf{Component} & \textbf{Specification} \\
\midrule
Input & $64\times480$ pixels, three channels \\
CNN backbone & 9 convolutional layers, channels $3{\rightarrow}64{\rightarrow}128{\rightarrow}256{\rightarrow}512$; each block $3\times3$ conv + batch norm + ReLU + 2D dropout (half rate); pooling $2\times2$, $2\times2$, then two $2\times1$ (height only), final unpadded $2\times2$; adaptive average pool over height $\rightarrow$ 512-d feature per sequence position \\
Sequence model & 4-layer bidirectional LSTM, $256$ hidden units per direction, inter-layer dropout $0.3$ $\rightarrow$ 512-d contextual features \\
Head & linear projection to character classes; CTC alignment; greedy decoding at inference \\
Parameters & $12{,}061{,}664$ trainable at the 32-symbol vocabulary ($5.74$M conv, $6.31$M recurrent, $16$K head) \\
\bottomrule
\end{tabular}
\end{table}

\begin{table}[htbp]
\caption{CRNN training configuration (identical across the four cells).}\label{tab:crnn_training}
\centering\small
\begin{tabular}{lp{0.66\linewidth}}
\toprule
\textbf{Item} & \textbf{Setting} \\
\midrule
Optimizer & AdamW, lr $1{\times}10^{-3}$, weight decay $0.05$, $\beta{=}(0.9,0.999)$, $\epsilon{=}10^{-8}$ \\
Schedule & CosineAnnealingWarmRestarts ($T_0{=}10$, $T_{\text{mult}}{=}2$, $\eta_{\min}{=}10^{-6}$), 5 warmup epochs \\
Duration & 100 epochs, batch size 16, single GPU; mixed precision; gradient clipping at max norm $1.0$ \\
Checkpoints & saved every epoch; all evaluated on SCI-val (Section~\ref{sec:methodology_selection}) \\
Random seed & config value $3407$ not applied by the training path; RNG state uncontrolled \\
Train transforms & resize $64\times480$; color jitter (brightness/contrast/saturation $0.1$, hue $0.05$); Gaussian blur ($k{=}3$, $\sigma\in[0.1,0.5]$, $p{=}0.1$); ImageNet normalization; Gaussian noise ($\sigma{=}0.01$, $p{=}0.1$). The same list is applied when computing validation loss during training \\
Inference path & deterministic: RGB cast, resize $64\times480$, tensor conversion, ImageNet normalization (used for all sweeps and reported metrics) \\
\bottomrule
\end{tabular}
\end{table}

\noindent The CRNN implementation inherited from
\citet{chung2025manchu} passes raw logits to PyTorch's
\texttt{CTCLoss}, which expects log-probabilities. As a robustness
check, we retrained all four CRNN cells with \texttt{log\_softmax}
added before the loss, with the remaining configuration identical and
run-to-run randomness not controlled. The main conclusions are
unaffected. Real-supervised training still far exceeds synthetic-only
training ($91.50$--$96.28\%$ versus $57.64\%$ ARCH-test WA), and the
CRNN still reaches the leading performance band. Only the fine-grained
ordering among the real-data regimes shifts, which is why the CRNN
composition comparisons in Section~\ref{sec:results} are stated for
the released recipe only.

\section{Full Results and Statistical Comparisons}\label{app:full_results}
Table~\ref{tab:perf_manchu} reports the word-accuracy and CER results
for all 16 selected configurations on all three splits.
Table~\ref{tab:paired_stats} reports the paired ARCH-test comparisons
cited in the main text.

\begin{table}[htbp]
\caption{Manchu-script performance of all 16 configurations on the complete
SYN-val split ($n{=}15{,}000$), the complete
SCI-val split ($n{=}3{,}359$;
Section~\ref{sec:methodology_selection}), and the
753-image archival ARCH-test set. Rows are ordered by
ARCH-test word accuracy. Each row corresponds to the
SCI-val-peak checkpoint selected per configuration
(Section~\ref{sec:methodology_selection}). WA: word accuracy;
CER: character error rate (micro-averaged, stripped). The 95\% CI column gives non-parametric bootstrap intervals for ARCH-test WA ($1{,}000$ resamples, percentile method).}\label{tab:perf_manchu}
\centering\small
\setlength{\tabcolsep}{5pt}
\begin{tabular}{l rr rr rrr}
\toprule
 & \multicolumn{2}{c}{SYN-val} & \multicolumn{2}{c}{SCI-val} & \multicolumn{3}{c}{ARCH-test} \\
\cmidrule(lr){2-3}\cmidrule(lr){4-5}\cmidrule(lr){6-8}
\textbf{Configuration} & WA & CER & WA & CER & WA & 95\% CI & CER \\
\midrule
LLaMA-SEQ           & 91.68 & 1.26 & 98.81 & 0.26 & \textbf{96.28} & [94.8, 97.6] & 0.78 \\
CRNN-REAL           & 67.66 & 7.32 & 99.32 & 0.14 & 95.88 & [94.4, 97.3] & 0.91 \\
CRNN-SEQ            & 76.91 & 4.76 & 99.14 & 0.25 & 95.75 & [94.3, 97.2] & 1.01 \\
Pixtral-SEQ         & 89.33 & 1.51 & 99.08 & 0.18 & 95.75 & [94.3, 97.2] & 0.96 \\
CRNN-JOINT          & 99.39 & 0.07 & 99.35 & 0.12 & 95.48 & [94.0, 96.9] & 0.88 \\
LLaMA-JOINT         & 97.91 & 0.28 & 98.18 & 0.41 & 95.09 & [93.5, 96.7] & 1.05 \\
Pixtral-JOINT       & 99.11 & 0.10 & 98.90 & 0.24 & 95.09 & [93.5, 96.7] & 1.13 \\
LLaMA-REAL          & 52.41 & 10.43 & 98.12 & 0.49 & 91.24 & [89.1, 93.2] & 2.28 \\
Pixtral-REAL        & 42.79 & 18.87 & 98.45 & 0.37 & 89.11 & [86.9, 91.4] & 2.62 \\
LLaMA-SYN           & 98.08 & 0.24 & 87.85 & 2.95 & 87.92 & [85.5, 90.2] & 3.43 \\
Qwen-SEQ            & 72.53 & 5.35 & 95.62 & 1.17 & 86.99 & [84.6, 89.4] & 4.19 \\
Qwen-JOINT          & 94.65 & 0.90 & 95.33 & 1.25 & 85.39 & [82.9, 87.9] & 4.07 \\
Pixtral-SYN         & 98.22 & 0.22 & 84.82 & 3.74 & 81.94 & [79.2, 84.6] & 4.85 \\
Qwen-REAL           & 15.10 & 31.54 & 92.20 & 2.63 & 70.92 & [67.7, 74.1] & 11.52 \\
CRNN-SYN            & 98.75 & 0.15 & 73.74 & 6.44 & 64.28 & [61.0, 67.9] & 10.62 \\
Qwen-SYN            & 93.65 & 1.05 & 69.78 & 8.83 & 62.42 & [58.8, 65.9] & 13.36 \\
\bottomrule
\end{tabular}
\end{table}

\begin{table}[htbp]
\caption{Paired comparisons on the shared 753-image ARCH-test
split. $\Delta$WA is the word-accuracy difference of the first-listed
system minus the second; $b/c$ counts the discordant items (first
correct / second wrong, and vice versa); the last column is a
$95\%$ paired bootstrap interval on $\Delta$WA ($20{,}000$ resamples
of the per-item difference vector). Ensemble rows use the
dictionary-guided ensemble of
Section~\ref{sec:results_ensemble}.}\label{tab:paired_stats}
\centering\small
\setlength{\tabcolsep}{5pt}
\begin{tabular}{l r c c}
\toprule
\textbf{Comparison} & $\Delta$WA (pp) & $b/c$ & 95\% paired CI (pp) \\
\midrule
LLaMA-SEQ $-$ CRNN-REAL & $+0.40$ & 20/17 & $[-1.20, +1.99]$ \\
CRNN-JOINT $-$ CRNN-SEQ & $-0.27$ & 21/23 & $[-1.99, +1.46]$ \\
CRNN-REAL $-$ CRNN-JOINT & $+0.40$ & 21/18 & $[-1.20, +1.99]$ \\
LLaMA-JOINT $-$ LLaMA-SEQ & $-1.20$ & 14/23 & $[-2.79, +0.40]$ \\
Pixtral-JOINT $-$ Pixtral-SEQ & $-0.66$ & 19/24 & $[-2.39, +1.06]$ \\
Qwen-JOINT $-$ Qwen-SEQ & $-1.59$ & 41/53 & $[-4.12, +0.93]$ \\
\midrule
QWJ ensemble $-$ best single (LLaMA-SEQ) & $+1.99$ & 16/1 & $[+0.93, +3.05]$ \\
QWJ ensemble $-$ majority vote & $+0.00$ & 2/2 & $[-0.53, +0.53]$ \\
\bottomrule
\end{tabular}
\end{table}

The selected-checkpoint results are sensitive to how tied SCI-val
peaks are resolved. Table~\ref{tab:tied_peaks} records the tied peaks
and the deterministic rule used to select among them. For CRNN-JOINT,
the selected checkpoint reaches $95.48\%$ ARCH-test WA, whereas the
other tied checkpoints reach $96.15$--$96.41\%$. The formal selection
rule remains fixed and the overall conclusion is unchanged, but this
spread should be considered when interpreting the CRNN JOINT--SEQ
comparison.

\begin{table}[htbp]
\caption{Tied SCI-val peaks and their resolution. All selected and
alternate checkpoints lie inside the leading band's confidence
intervals.}\label{tab:tied_peaks}
\centering\small
\begin{tabular}{llll}
\toprule
\textbf{Cell} & \textbf{Tie} & \textbf{Decided by} & \textbf{Test WA sel. / alt.} \\
\midrule
Pixtral-REAL & 2 at $98.45\%$ & lower CER & $89.11$ / --- \\
CRNN-SEQ & 3 at $99.14\%$ & lower CER & $95.75$ / $95.88$--$96.02$ \\
CRNN-JOINT & 3 at $99.35\%$ & lower CER (6th decimal) & $95.48$ / $96.15$--$96.41$ \\
Pixtral-SEQ & 2 at $99.08\%$ & earlier step (CER also tied) & $95.75$ / --- \\
\bottomrule
\end{tabular}
\end{table}

\end{document}